\documentclass[letterpaper]{article}
\usepackage{aaai}
\usepackage{times}
\usepackage{helvet}
\usepackage{courier}
\usepackage{graphicx}
\usepackage{color}
\usepackage[tight,footnotesize]{subfigure}
\usepackage[cmex10]{amsmath}
\begin{document}
%
\title{Feature Enhancement Network: A Refined Scene Text Detector}
\author{Sheng Zhang, Yuliang Liu, Lianwen Jin, Canjie Luo\\
School of Electronic and Information Engineering, South China University of Technology\\
zsscut90@gmail.com, liu.yuliang@mail.scut.edu.cn, \{lianwen.jin, canjie.luo\}@gmail.com\\
}
\maketitle
\begin{abstract}
\begin{quote}
In this paper, we propose a refined scene text detector with a \textit{novel} Feature Enhancement Network (FEN) for Region Proposal and Text Detection Refinement. Retrospectively, both region proposal with \textit{only} $3\times 3$ sliding-window feature and text detection refinement with \textit{single scale} high level feature are insufficient, especially for smaller scene text. Therefore, we design a new FEN network with \textit{task-specific}, \textit{low} and \textit{high} level semantic features fusion to improve the performance of text detection. Besides, since \textit{unitary} position-sensitive RoI pooling in general object detection is unreasonable for variable text regions, an \textit{adaptively weighted} position-sensitive RoI pooling layer is devised for further enhancing the detecting accuracy. To tackle the \textit{sample-imbalance} problem during the refinement stage, we also propose an effective \textit{positives mining} strategy for efficiently training our network. Experiments on ICDAR 2011 and 2013 robust text detection benchmarks demonstrate that our method can achieve state-of-the-art results, outperforming all reported methods in terms of F-measure.
\end{quote}
\end{abstract}

\section{Introduction}
Text detection in natural scene is an important component (\cite{bissacco2013photoocr}, \cite{yin2014robust}, \cite{ye2015text}, \cite{He2016Text}, \cite{he2017single}) for various intelligent applications based on computer vision. For instance, blind navigation, multilingual translation, automotive assistance and image-based geolocation, etc. Different from conventional OCR technique, scene text detection are often challenged by perspective distortions, variation of text size, color or uncontrollable illumination intensity, etc.

In recent years, various methods have been successfully applied for detecting scene text. However, they usually comprise several processing steps, e.g. character or word proposal generation (\cite{neumann2012real}, \cite{huang2014robust}, \cite{jaderberg2016reading}), proposals filtering and clustering. They often entail much effort in designing heuristic rules and tuning parameters to make each module conduct well, which conversely reduces the speed of text detection. 

Currently, Deep Convolutional Neural Networks (DCNN) have advanced general object detection substantially, which is also widely used in the scene text detection field. The main differences between general object detection and scene text detection have been discussed by some previous published papers such as (\cite{tian2015text}, \cite{sun2015robust}), which includes that the scene text are often smaller, thinner and with rich diversity on the aspect-ratios and so on, which makes the detection of scene text a very challenging problem different from general object detection. Besides, for a better performance, almost all object detection algorithms which adopt the DCNN framework take the strategy: i.e. training the proposed network by fine-tuning a model pre-trained on ImageNet dataset \cite{russakovsky2015imagenet} which is used for image recognition, a different task which requires the extracted features are \textbf{position-insensitive}. Contradictorily, the task for object detection is \textbf{position-sensitive}. Therefore, \cite{dai2016r} proposes the \textit{position-sensitive} RoI pooling layer to solve the problem. However, they \textit{only} use the $ 3\times 3$ sliding-window feature for region proposal and \textit{single scale} high level feature for the refinement of object detection, which is insufficient for general text detection task, specially for smaller text regions; Meanwhile, their \textit{unitary} position-sensitive RoI pooling in general object detection is unreasonable for much variable text regions. Motivated by their work, we propose a refined scene text detection framework via a \textit{novel} Feature Enhancement Network (FEN)  which can directly generate word bounding boxes and be end-to-end trainable.

Our key contributions in this paper are as follows:
\begin{itemize}
\item We propose a novel FEN network which promotes the recall rate and accuracy of text detection.
\item We present an \textit{adaptively weighted} position-sensitive Region of Interest (RoI) pooling module to further raise the accuarcy of the Text Detection Refinement stage.
\item We propose a \textit{positives mining} strategy to solve the \textit{sample imbalence} problem during the Text Detection Refinement stage.
\item Our approach shows state-of-the-art performance on ICDAR 2011 and 2013 robust text detection benchmarks.
\end{itemize}

\begin{figure*}[t]
\begin{center}
\includegraphics[width=0.98\textwidth]{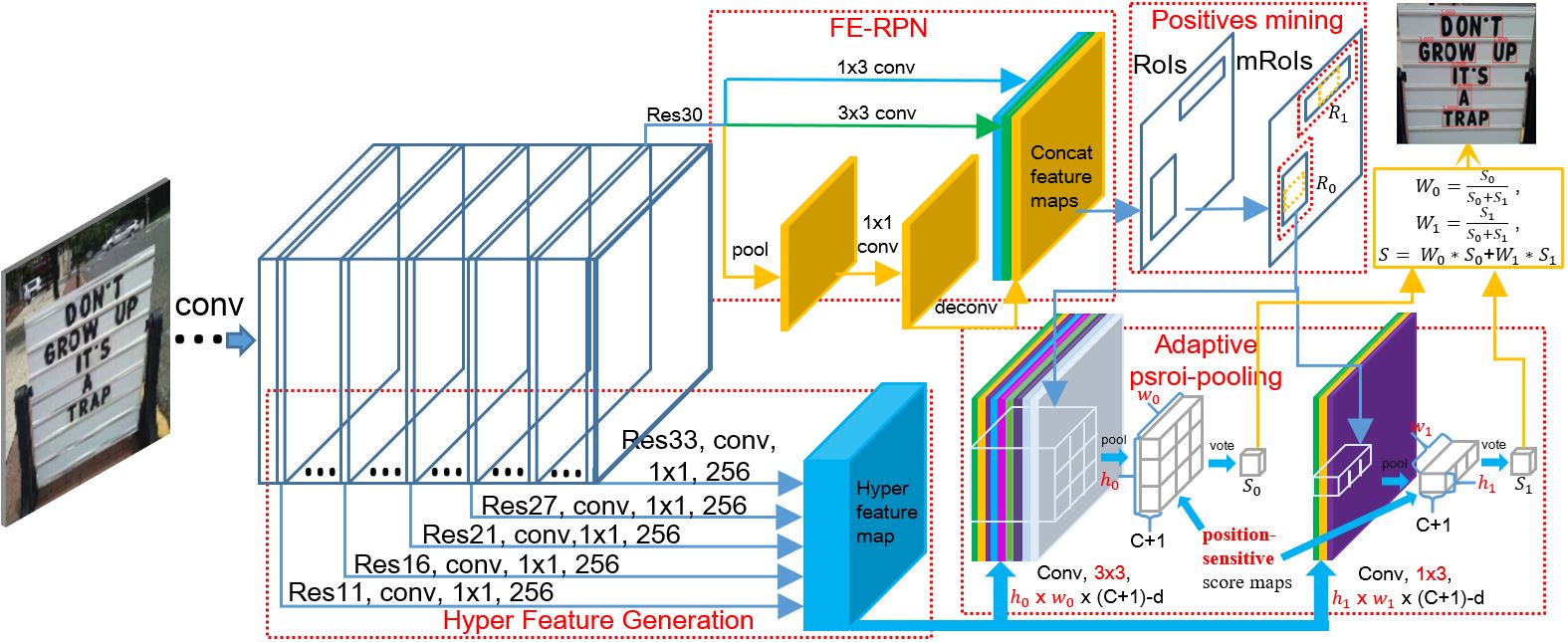}
\caption{The overall architecture of our FEN. It consists of three innovative components. 1, Feature Enhancement network stem with Feature Enhancement RPN (FE-RPN) and Hyper Feature Generation; 2, Positives mining; 3, Adaptively weighted position-sensitive RoI pooling.}\label{fig:quaR}
\end{center}
\end{figure*}

\section{Related Works}
In general, text spotting in wild scene can be detailedly categorized by two sub-tasks: text detection and text recognition. The former focuses on accurate text localization in terms of word or text-line bounding boxes in scene images; The latter translates the localized text regions into machine-interpretable character sequences. In this paper, we only concentrate on the text detection task.

Currently, we roughly group mainstream methods for text detection into three categories: \\
$\left(1\right)$ \textbf{Character-based:} this kind of method mainly concentrates on the characteristics of individual characters and the relationships between characters \cite{zhu2016scene}, \textit{e.g.} connected component based methods (\cite{zamberletti2014text}, \cite{shi2013scene}). Some of existed approaches often take advantage of Stroke Width Transform (SWT) \cite{huang2013text} or Maximally Stable Extremal Region (MSER) \cite{nister2008linear} algorithms to first extract character candidates and then use successive steps to filter non-text patches for exactly connecting the candidates. Although such methods are accurate, they are partly constrained to preserve various true character patches in practice \cite{cho2016canny}.\\
$\left(2\right)$ \textbf{Word-based:} Text words are considered as the general object to be detected (\cite{zhong2016deeptext}, \cite{gomez2017textproposals}). \cite{zhong2016deeptext} proposes a Faster-RCNN based method \cite{ren2015faster}. Word proposals are generated with Region Proposal Network and then followed by an embeded text detection refinement stage. However, their performance is not perfect. \cite{gupta2016synthetic} adopts the YOLO framework \cite{redmon2016you} which directly regresses the text localizations on many regular grids and runs very fast. Nevertheless, it does not perform well on a group of small text regions.\\
$\left(3\right)$ \textbf{Text-line-based:} Text lines are first detected by ignoring the noise of blank space between words and then partitioned into individual words. For instance, \cite{zhang2015symmetry} leverages the symmetric characteristics of text regions to detect text lines and further puts forward to detect text lines with fully convolutional networks \cite{long2015fully} in year 2016 \cite{zhang2016multi}.

Our method is inspired by R-FCN \cite{dai2016r} and on the basis of text word. Different from original R-FCN network, we have improved the network by \textit{task-specific}, \textit{low} and \textit{high} level semantic features fusion, which observably promote the performance of text detection. Besides, we propose a \textit{positives mining} strategy and an \textit{adaptively weighted} position-sensitive Region of Interest pooling layer which can both remarkably improve the accuarcy of text detection.

\section{Proposed Methodology}
The overall architecture of our FEN network is elaborated in Figure \ref{fig:quaR}. To the best of our knowledge, mainstream methods for general object detection often comprise two stages: proposals generation and detection refinement, so is the R-FCN \cite{dai2016r} framework. In this paper, we integrate three \textit{innovative} components into the R-FCN framework which inherits the well-known ResNet-101 architecture \cite{he2016deep} and removes the last global pooling, classification layers. By the FEN network, we first enhance the text feature for region proposal through the \textit{FE-RPN} and for text detection refinement with Hyper Feature Generation module; Then, we use \textit{Positives mining strategy} on the region proposals to adjust the ratio between positive and negative samples; Finally, during the text detection refinement stage, we apply the \textit{adaptively weighted} position-sensitive RoI pooling on the hyper-features to produce the accurate text detection results.

\subsection{Feature Enhancement Network Stem}
In our FEN network, we use the ResNet-101 \cite{he2016deep} network as our backbone network. Besides, we have integrated following two components into it.\\
$1)$ \textbf{Feature Enhancement RPN (FE-RPN):}
Before, researchers in the field of general object detection always generate region proposals with \textit{only} the $3\times 3$ convolutional sliding-window feature on some intermediate layer, which we reckon that it is inadequate for text region proposal. Since widths of most words or text lines are generally much greater than heights; In other words, their aspect-ratios are usually much greater than one. Meanwhile, high level semantic feature has much larger receptive field and contains much contextual information which is conducive to distinguish foreground objects from background objects. As is shown in Figure 1 top-center: FE-RPN, we select \textit{Res30} layer from the network stem as the input layer of FE-RPN and add two branches, one is the text-characteristic and task-specific $1\times 3$ convolution layer, the other holds max pooling, $1\times 1$ convolution, deconvolution layers, which has two superiorities: 1) what we used is the deconvolution layer proposed by \cite{odena2016deconvolution} which can eliminate the influence of checkerboard artifacts of conventional deconvolution operation radically, 2) make the salient feature more salient and extract much contextual information; Then, we concatenate the original $3\times 3$ convolutional sliding-window feature with the outputs of deconvolution layer and $1\times 3$ convolution layer; Finally, a new convolution layer as well as a \textit{ResNet block} are exploited to achieve the goal of feature enhancement for region proposal.
 \\
 \\
$2)$ \textbf{Hyper Feature Generation:} Previous object detection approaches always make full use of \textit{single scale} and \textit{high level} semantic feature to conduct the refinement of object detection, which may lose much information of object details and thus insufficient for accurate objection localization, especially for smaller text regions. However, the low level semantic feature from intermediate layers will make compensate for the high level semantic feature because of its capability of detail retention. In a word, \textit{high level} semantic feature is conducive to object classification while \textit{low level} feature is beneficial for accurate object localization. Accordingly, \cite{kong2016hypernet} propose the HyperNet to reinforce the feature map for accurate object localization. In their HyperNet, feature maps originated from different intermediate layers have different \textit{spatial size} and are merged together by pooling, convolution, deconvolution operations, which is computational complexity. For the simplicity, we just compress the intermediate feature maps along the channel dimension with \textit{bottleneck} convolution, which is widely acknowledged to refine the salient features effectively. And more importantly, the intermediate feature maps which we reuse for the hyper-feature are \textit{originally} of the \textit{same} spatial size, which has three superiorities: 1) directly provide the deep supervision information for some intermediate residual blocks; 2) parallelize the residual learning rather than the conventional serialization residual learning; 3) improve the efficiency of computation simultaneously (see the left-bottom of Figure \ref{fig:quaR}).
 
\subsection{Text Proposals Generation}
$1)$ \textbf{Text characteristic anchor design:} Anchor mechanism \cite{ren2015faster} is yet the most classical mechanism for accurate object detection, and different tasks should have different anchor design principle. Specific text detection task also has text characteristic anchor design. As aforementioned, aspect ratios of most words or text lines are prone to be larger than one. Besides, text regions are normally smaller than other general objects in natural scene. We empirically select six scales (32, 64, 112, 192, 304 and 416) and five aspect ratios (1, 2, 3, 4 and 6). Nevertheless, some anchors designed before may be useless and unreasonable. For example, $scale = 416, aspect$\_$ratio = 6$, it is almost impossible that a text word takes over the whole input image. Finally, we manually keep a total of $a = 24$ anchors at each feature point of the feature enhancement map from FE-RPN sub-network.

Subsequently, we impose the well-designed anchors on each feature point of the feature enhancement map by Equation \ref{decode} to attain candidate word proposals.
\begin{equation}\label{decode}
\begin{split}
  &x = x_0 + w_0 \times \Delta x, \\
  &y = y_0 + h_0 \times \Delta y, \\
  &w = w_0 \times exp\left(\Delta w \right),\\
  &h = h_0 \times exp\left(\Delta h \right)
\end{split}
\end{equation}
Where $\left(x_0, y_0, w_0, h_0\right)$ represents the center coordinate, width, height of each anchor, $\left(x, y, w, h\right)$ indicates the predicted center coordinate, width, height of each proposal, $\left(\Delta x, \Delta y, \Delta w, \Delta h\right)$ is the output of our FE-RPN sub-network and simultaneously denotes the predicted offset for each proposal relate to corresponding anchor. Yet, numerous candidate word proposals are surplus and highly overlap with each other. For this reason, we apply NMS \cite{neubeck2006efficient} with an Intersection over Union (IoU) overlap threshold of 0.7 to candidate word proposals to hold the first 200 proposals which are ranked by the scores from FE-RPN sub-network. The held 200 proposals most possibly cover all text regions emerged in the input images and will be fed into the following \textit{Positives mining} layer.\\
 \\
$2)$ \textbf{Positives Mining:} Basically, the batchsize of input images in R-FCN \cite{dai2016r} framework is \textit{only one}, which owns two traits: enhancing the memory of each emerged sample; leading to \textit{samples imbalance}, especially that input images in ICDAR 2011 and 2013 robust text detection datasets frequently contain \textit{only one} text sample and the text region is small, which will cover few anchors for training. Hence, we opt the Positives Mining strategy.
\begin{figure}[!htbp]
\begin{center}
\includegraphics[width=0.40\textwidth]{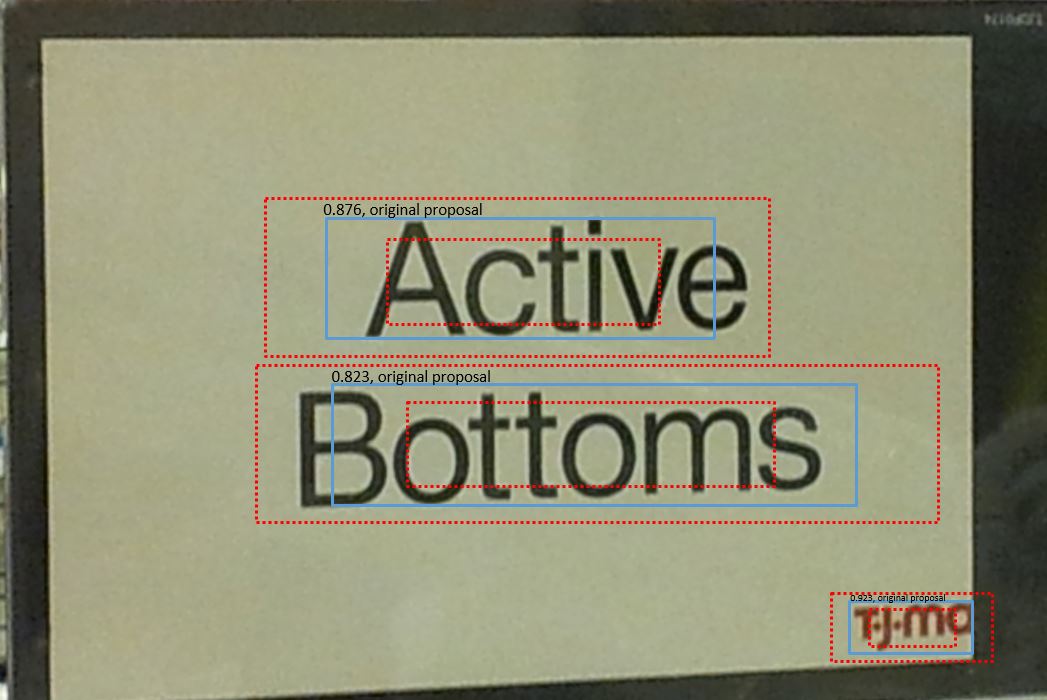}
\caption{Positives mining. Boxes with \textit{shallow blue} color are original proposals, others are scaled proposals.}\label{fig:pos}
\end{center}
\end{figure}
\noindent As Figure \ref{fig:pos} shows, boxes with \textit{shallow blue} color and textness scores are original proposals generated from \textit{FE-RPN} sub-network, others are scaled proposals with scales $\{0.7, 1.3\}$. With these scales, the IoU overlap between the new proposals and original proposals will not exceed previously acknowledged threshold of the RPN NMS $(i.e.\ $= 0.7$)$. Since we observe that almost all positive proposals rank in top 50 by their scores and we \textit{only} select the first fifty original proposals as references, we call it \textit{Positives mining} which can obviously solve the problem of \textit{samples imbalance}.

\subsection{Text Detection Refinement}
Although the text proposal generation stage can recall almost all text regions, it will have a much lower precision for text detection. Thus, in a similar way, we also adopt the text detection refinement stage, which is innovative and different from \cite{dai2016r} framework.\\
$1)$ \textbf{Adaptively Weighted Position-Sensitive RoI Pooling:} In \cite{dai2016r} framework, they partition each RoI into $k \times k$ $\left(k=7\right)$ bins by a regular grid to encode position information. For an RoI with size $w \times h$, each bin is of size $\frac{w}{k} \times \frac{h}{k}$. Then, they produce $k^2$ score maps for each category by the last convolutional layer. Inside the $\left(i, j\right)$-th bin $(0 < i, j < k-1)$, they compute the value of \textit{position-sensitive} RoI pooling layer by pooling \textit{only} over the corresponding position in the $(i, j)$-th score map:
\begin{equation}\label{psroi}
r_c\left(i,j|\Theta\right)=\sum\limits_{\left(x,y\right)\in bin\left(i,j\right)}{\frac{z_{i,j,c}\left(x + x_0, y + y_0|\Theta\right)}{n}}
\end{equation}
In Equation \ref{psroi}, $r_c\left(i,j|\Theta\right)$ is the pooled value in the $(i,j)$-th bin for category $c$, $z_{i,j,c}$ represents a score map from the $k^2(C + 1)$ score maps. $(x_0, y_0)$ indicates the top-left corner coordinate of an RoI, $n$ denotes the amount of pixels in the bin, and $\Theta$ stands for all network parameters. After the \textit{position-sensitive} RoI pooling procedure, they will get the \textit{coarse} scores of each RoI via globally pooling on the $k^2$ \textit{position-sensitive} score maps: $r_c\left(\Theta\right)=\sum{\frac{r_c\left(i,j|\Theta\right)}{k^2}}$, which produces a $(C + 1)$-dimensional vector. 
Finally, with the softmax operation across all categories:
\begin{equation}\label{soft}
s_c\left(\Theta\right)=\left.{e^{r_c\left(\Theta\right)}}\middle/\right. {\sum\limits_{c'=0}^C{e^{r_{c'}\left(\Theta\right)}}}
\end{equation}
They obtain the \textit{expected} scores which are used for calculating the cross-entropy loss during training and for ranking the RoIs during inference.

As for the bounding box regression, it will be done in a similar way except replacing the aforementioned $k^2(C+1)$-d
convolution layer with a sibling, class-agnostic $4k^2$-d convolution layer.

\begin{figure}[h]
\begin{center}
\includegraphics[width=0.3\textwidth]{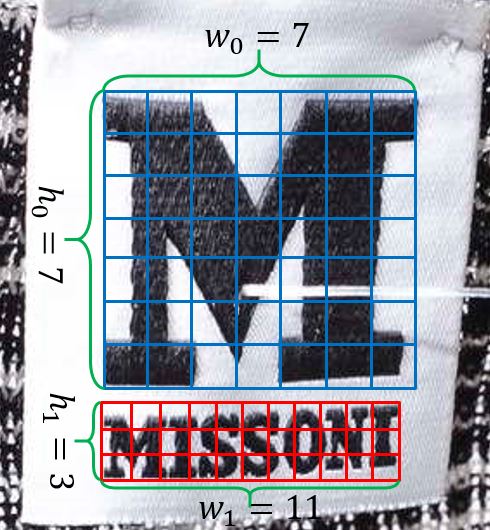}
\caption{Practical position-sensitive RoI pooling}\label{fig:prac}
\end{center}
\end{figure}

Although their \textit{position-sensitive} RoI pooling layer performs well on general object detection task, it behaves badly on the scent text detection task. Not only because text regions are normally \textit{smaller} than general objects but also owing to the \textit{exaggerated} aspect-ratios of text words. For example, two text samples with spatial sizes $48\times48$, $48\times256$ are in an input image, after the \textit{forward-propagate} procedure, the corresponding text regions in the feature map \textit{Res30} will be $3\times3$, $3\times16$ respectively, and now it is obviously unreasonable to impose the \textit{position-sensitive} RoI pooling operation with \textit{conventional} size $7\times7$. On the contrary, the \textit{practical} \textit{position-sensitive} RoI pooling should be as Figure \ref{fig:prac}.
Accordingly, as Figure \ref{fig:quaR} shows, we put forward the \textit{Adaptively Weighted} Position-Sensitive RoI Pooling layer to solve the problem. Firstly, we divide each RoI into $w_l \times h_l$ bins, $\left(w_l\times h_l\right)\in\left\{3\times 3, 7\times 7, 3\times 8, 3\times 11 | l=0,1,2,3\right\}$; Secondly, with the previously generated hyper-feature maps, we produce four kinds of $w_l \times h_l$ score maps with \textit{text-characteristic} filter sizes for each category; then, after applying position-sensitive RoI pooling and global pooling for each RoI to the four kinds of score maps, each RoI will correspond to four scores $\left(S_l\left(\Theta\right)|l=0,1,2,3\right)$, which represent the textness scores with adaptive pooling sizes $\left\{3\times 3, 7\times 7, 3\times 8, 3\times 11\right\}$; Finally, we will leverage the four scores to evaluate the adaptive weight $W_l\left(\Theta\right)$ for each kind of pooling size and the final textness score $S\left(\Theta\right)$ is as follow:
\begin{equation}\label{ada_w}
W_l\left(\Theta\right)=\frac{S_l\left(\Theta\right)}{\sum\limits_{l'=0}^4{S_{l'}\left(\Theta\right)}}, \ \  \left(l=0,1,2,3\right)
\end{equation}
\begin{equation}\label{ada_s}
S\left(\Theta\right)=S_l\left(\Theta\right)\times {W_l\left(\Theta\right)}, \ \  \left(l=0,1,2,3\right)
\end{equation}
Clearly, different pooling sizes are suitable for different text regions which own different spatial sizes and aspect-ratios, the most suitable pooling size will get the highest score. Correspondingly, its adaptive weight will be the highest and it will naturally contribute most to the final textness score. Vice versa. Moreover, with regard to bounding-box regression, we will share the evaluated adaptive weight and do it in the same way, i.e.
\begin{equation}\label{ada_b}
B\left(\Theta\right)=B_l\left(\Theta\right)\times {W_l\left(\Theta\right)}, \ \  \left(l=0,1,2,3\right)
\end{equation}
which makes our approach achieve the state-of-the-art results.

\begin{figure*}[!htp]
\centering
\subfigure[R: 100\%; P: 100\%; H: 100\%]{
\label{Fig.sub.1}
\includegraphics[width=0.28\textwidth, height=0.14\textwidth]{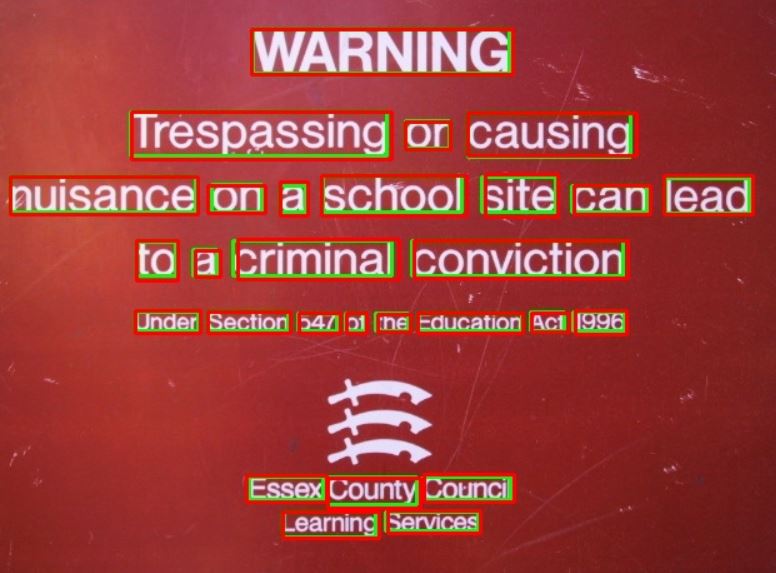}}
\hspace{-2mm}
\subfigure[R: 100\%; P: 100\%; H: 100\%]{
\label{Fig.sub.2}
\includegraphics[width=0.28\textwidth, height=0.14\textwidth]{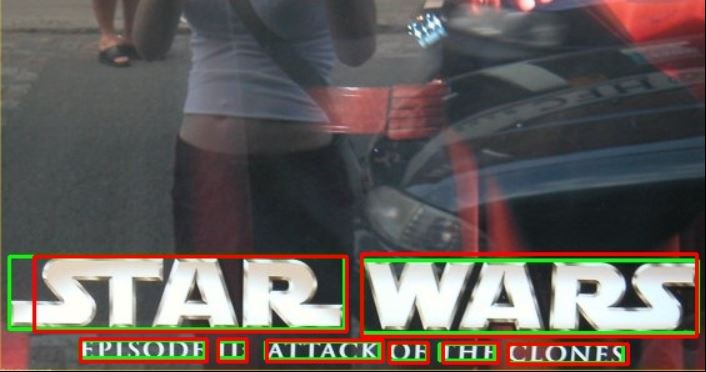}}
\hspace{-2mm}
\subfigure[R: 100\%; P: 100\%; H: 100\%]{
\label{Fig.sub.3}
\includegraphics[width=0.28\textwidth, height=0.14\textwidth]{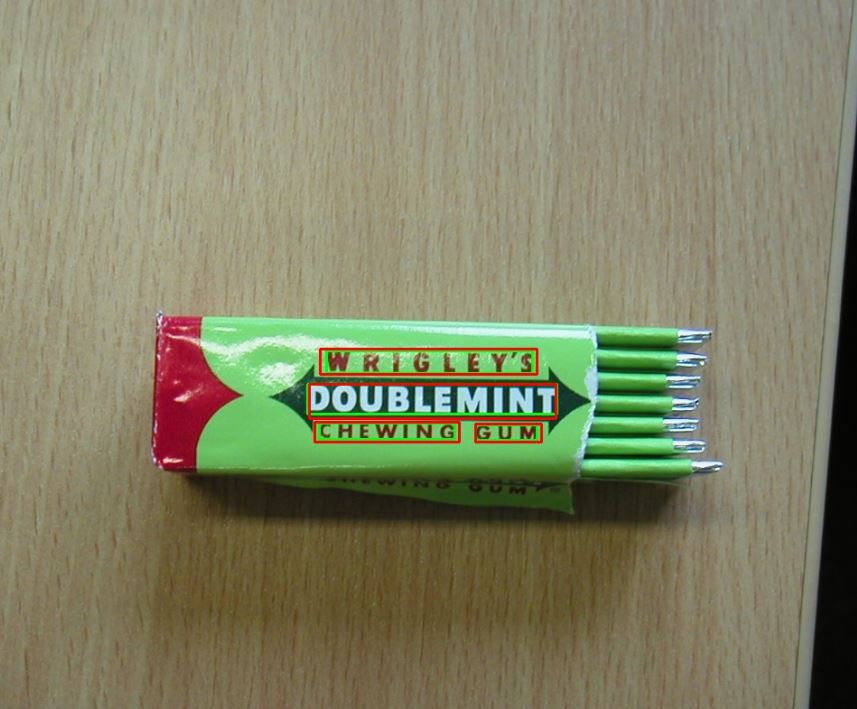}}
\vfill
\subfigure[R: 100\%; P: 100\%; H: 100\%]{
\label{Fig.sub.4}
\includegraphics[width=0.28\textwidth, height=0.14\textwidth]{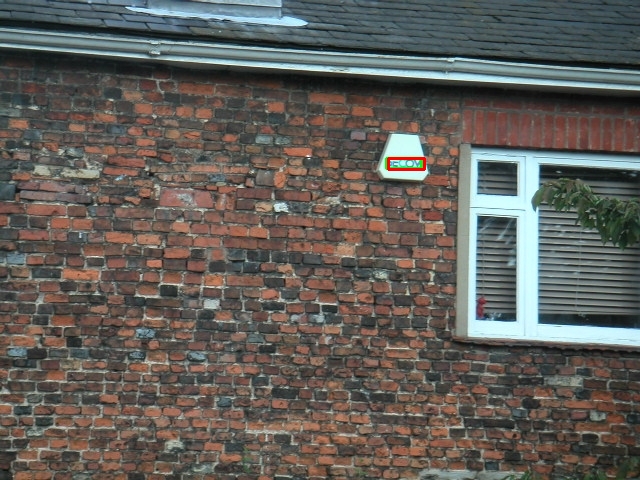}}
\hspace{-2.1mm}
\subfigure[R: 100\%; P: 100\%; H: 100\%]{
\label{Fig.sub.5}
\includegraphics[width=0.28\textwidth, height=0.14\textwidth]{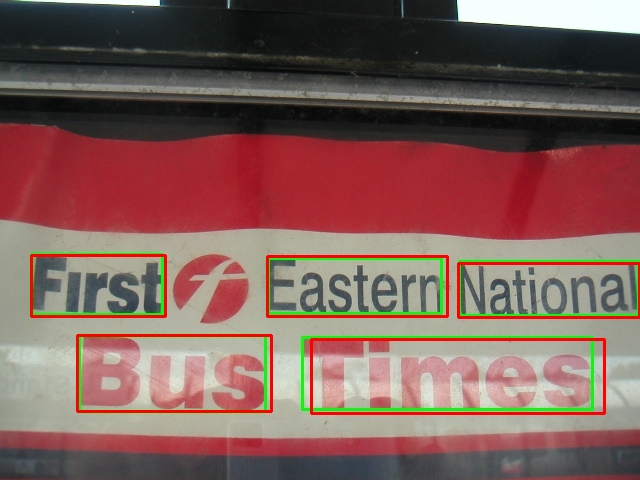}}
\hspace{-2.1mm}
\subfigure[R: 100\%; P: 100\%; H: 100\%]{
\label{Fig.sub.6}
\includegraphics[width=0.28\textwidth, height=0.14\textwidth]{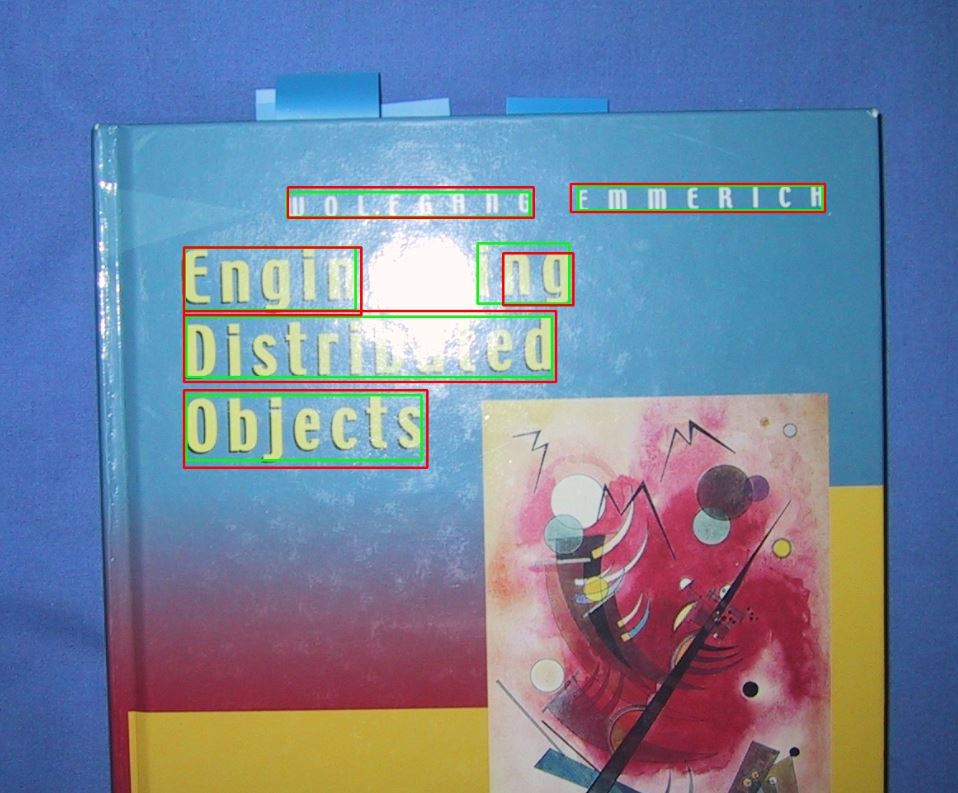}}
\vfill
\subfigure[R: 98.33\%; P: 96.92\%; H: 97.62\%]{
\label{Fig.sub.7}
\includegraphics[width=0.28\textwidth, height=0.14\textwidth]{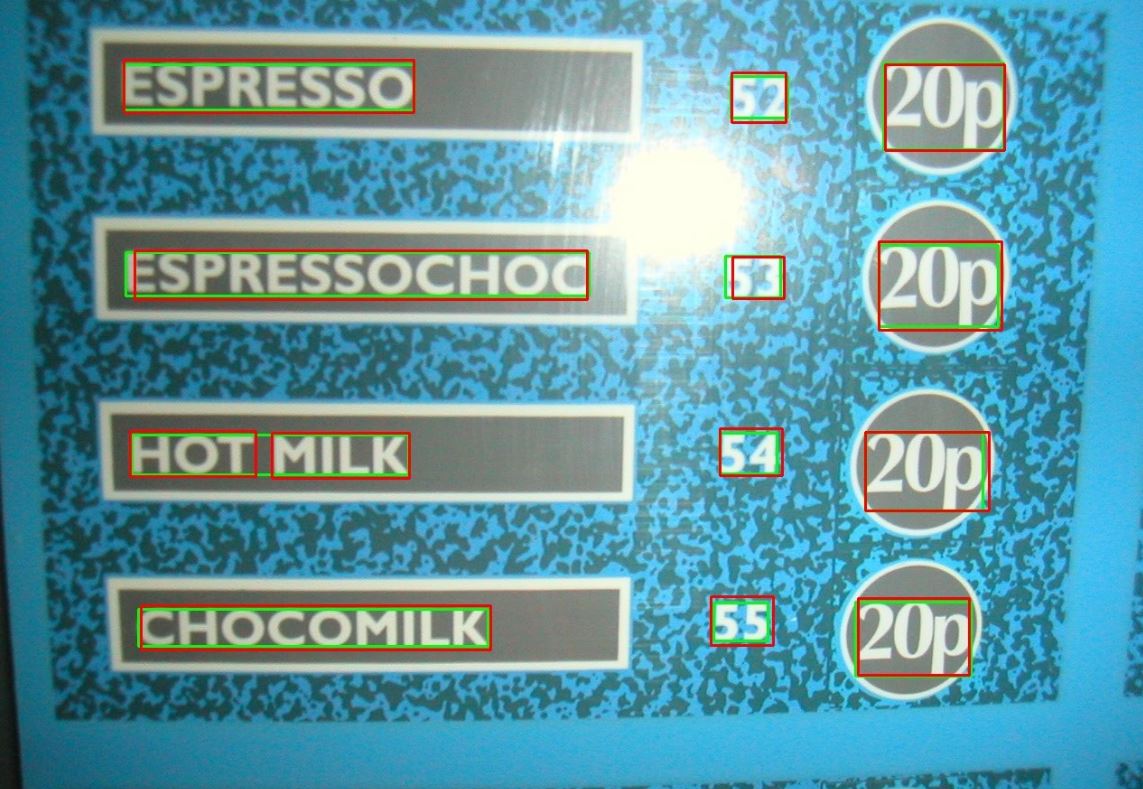}}
\hspace{-2.1mm}
\subfigure[R: 100\%; P: 85.71\%; H: 92.31\%]{
\label{Fig.sub.8}
\includegraphics[width=0.28\textwidth, height=0.14\textwidth]{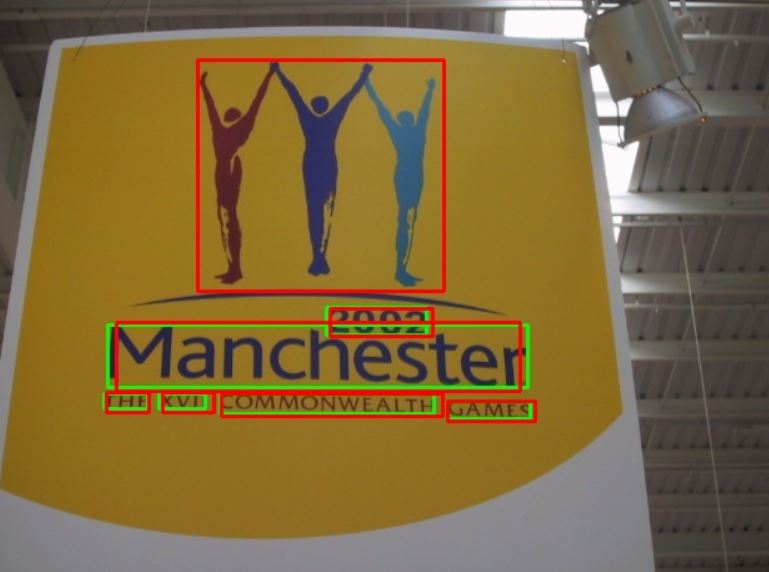}}
\hspace{-2.1mm}
\subfigure[R: 66.67\%; P: 80\%; H: 72.73\%]{
\label{Fig.sub.9}
\includegraphics[width=0.28\textwidth, height=0.1\textwidth]{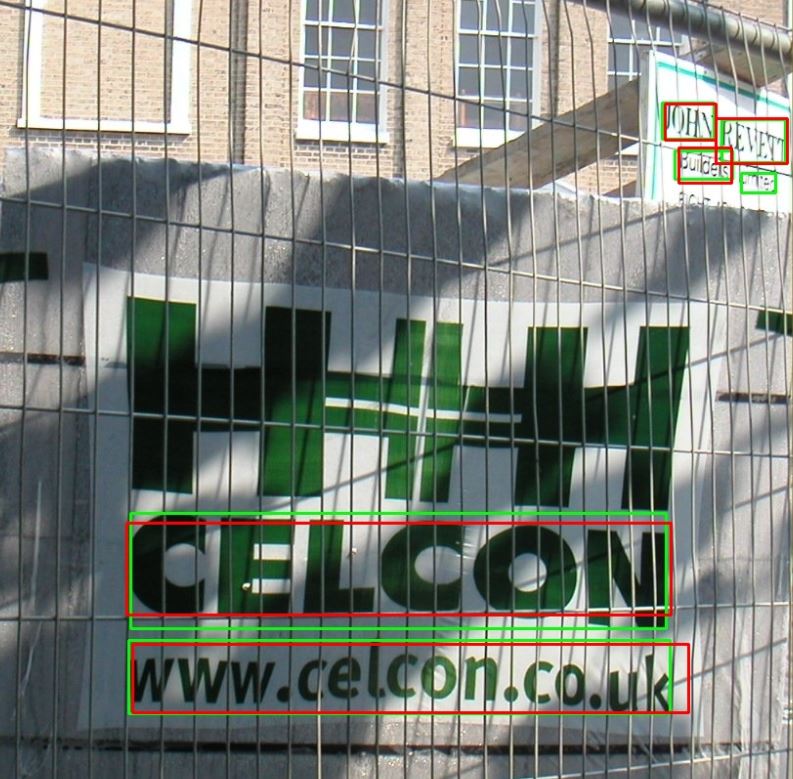}}
\caption{Qualitative evaluation results on ICDAR datasets. Green boxes are ground-truth boxes, red boxes are detected results. R: recall; P: precision; H: H-mean (F-measure)}
\label{Fig.label}
\end{figure*}

\begin{table*}[!htp]
\caption{The effectiveness of different components of our method on ICDAR 2011 and 2013 robust text detection datasets. IC13 Eval: ICDAR 2013 evaluation criterion; DetEval: \cite{wolf2006object}; R: recall; P: precision; F: F-measure. PM: Positives Mining. FENS: Feature Enhancement Network Stem. MT: multi-scale test.}
\scriptsize
\begin{center}
\begin{tabular}{|c|c|c|c|c|c|c|c|c|c|c|c|c|c|c|c|c|}
\hline
\multicolumn{4}{|c|}{Datasets} & \multicolumn{6}{|c|}{ICDAR 2011} & \multicolumn{6}{|c|}{ICDAR 2013} & Time\slash s\\
\hline
\multicolumn{4}{|c|}{Evaluation protocol} & \multicolumn{3}{|c|}{IC13 Eval} & \multicolumn{3}{|c|}{DetEval} & \multicolumn{3}{|c|}{IC13 Eval} & \multicolumn{3}{|c|}{DetEval} & \\
\hline
\multicolumn{4}{|c|}{Methods} & R & P & F & R & P & F & R & P & F & R & P & F & \\
\hline
\multicolumn{4}{|c|}{R-FCN \cite{dai2016r}} & 0.812 & 0.873 & 0.841 & 0.824 & 0.881 & 0.852 & 0.819 & 0.907 & 0.861 & 0.827 & 0.910 & 0.867 & \textbf{0.14}\\
\hline
\multicolumn{4}{|c|}{R-FCN + our PM} & 0.810 & 0.895 & 0.850 & 0.821 & \textbf{0.899} & 0.858 & 0.821 & 0.918 & 0.867 & 0.825 & 0.923 & 0.871 & 0.16\\
\hline
\multicolumn{4}{|c|}{R-FCN + our FENS} & 0.865 & 0.874 & 0.869 & 0.872 & 0.886 & 0.879 & 0.878 & 0.907 & 0.892 & 0.884 & 0.918 & 0.90 & 0.27\\
\hline
\multicolumn{4}{|c|}{our FEN} & 0.875 & 0.892 & 0.884 & 0.881 & 0.894 & 0.887 & 0.891 & 0.936 & 0.913 & 0.897 & 0.939 & 0.918 & 0.31\\
\hline
\multicolumn{4}{|c|}{our FEN + MT} & \textbf{0.889} & \textbf{0.896} & \textbf{0.892} & \textbf{0.895} & 0.898 & \textbf{0.897} & \textbf{0.893} & \textbf{0.941} & \textbf{0.916} & \textbf{0.90} & \textbf{0.947} & \textbf{0.923} & 0.90\\
\hline
\end{tabular}
\end{center}
\label{tab:self}
\end{table*}

\begin{table*}[!htp]
\caption{Comparison with state-of-the-art methods on ICDAR 2011 and 2013 robust text detection datasets. IC13 Eval: ICDAR 2013 evaluation criterion; DetEval: \cite{wolf2006object}; R: recall; P: precision; F: F-measure. MT: multi-scale test.}
\newcommand{\tabincell}[2]{\begin{tabular}{@{}#1@{}}#2\end{tabular}}
\scriptsize
\begin{center}
\begin{tabular}{|c|c|c|c|c|c|c|c|c|c|c|c|c|c|c|c|c|}
\hline
\multicolumn{4}{|c|}{Datasets} & \multicolumn{6}{|c|}{ICDAR 2011} & \multicolumn{6}{|c|}{ICDAR 2013} & Time\slash s\\
\hline
\multicolumn{4}{|c|}{Evaluation protocol} & \multicolumn{3}{|c|}{IC13 Eval} & \multicolumn{3}{|c|}{DetEval} & \multicolumn{3}{|c|}{IC13 Eval} & \multicolumn{3}{|c|}{DetEval} & \\
\hline
\multicolumn{4}{|c|}{Methods} & R & P & F & R & P & F & R & P & F & R & P & F & \\
\hline
\multicolumn{4}{|c|}{TextFlow \cite{tian2015text}} & 0.76 & 0.86 & 0.81 & - & - & - & 0.76 & 0.85 & 0.80 & - & - & - & 1.4\\
\hline
\multicolumn{4}{|c|}{\tabincell{c}{FCRNall + filts \\ \cite{gupta2016synthetic}}} & - & - & - & 0.75 & 0.92 & 0.82 & - & - & - & 0.76 & 0.92 & 0.83 & $>$1.27\\
\hline
\multicolumn{4}{|c|}{FCN \cite{zhang2016multi}} & - & - & - & - & - & - & 0.78 & 0.88 & 0.83 & - & - & - & 2.1\\
\hline
\multicolumn{4}{|c|}{TextBoxes + MT \cite{liao2017textboxes}} & 0.82 & 0.88 & 0.85 & 0.82 & 0.89 & 0.86 & 0.83 & 0.88 & 0.85 & 0.83 & 0.89 & 0.86 & 0.73\\
\hline
\multicolumn{4}{|c|}{DeepText \cite{zhong2016deeptext}} & - & - & - & 0.81 & 0.85 & 0.83 & 0.826 & 0.904 & 0.863 & 0.842 & 0.907 & 0.873 & 1.7\\
\hline
\multicolumn{4}{|c|}{CTPN \cite{tian2016detecting}} & - & - & - & 0.79 & 0.89 & 0.84 & 0.83 & 0.93 & 0.88 & - & - & - & \textbf{0.14}\\
\hline
\multicolumn{4}{|c|}{EASTtext \cite{zhou2017east}} & - & - & - & - & - & - & 0.827 & 0.926 & 0.874 & - & - & - & 0.75\\
\hline
\multicolumn{4}{|c|}{our FEN} & 0.875 & 0.892 & 0.884 & 0.881 & 0.894 & 0.887 & 0.891 & 0.936 & 0.913 & 0.897 & 0.939 & 0.918 & 0.31\\
\hline
\multicolumn{4}{|c|}{our FEN + MT} & \textbf{0.889} & \textbf{0.896} & \textbf{0.892} & \textbf{0.895} & \textbf{0.898} & \textbf{0.897} & \textbf{0.893} & \textbf{0.941} & \textbf{0.916} & \textbf{0.90} & \textbf{0.947} & \textbf{0.923} & 0.90\\
\hline
\end{tabular}
\end{center}
\label{tab:ic11}
\end{table*}

\subsection{Optimization}
During the training procedure, we choose the similar multi-task loss functions for both text region proposal and text detection refinement stages, i.e. 
\begin{equation}\label{loss}
L\left(s,c,b,g\right)=\frac{1}{N}\left(L_{cls}{\left(s,c\right)} + \lambda\times L_{loc}{\left(b,g\right)}\right)
\end{equation}
where $N$ is the amount of anchors or proposals that match ground-truth boxes, and $\lambda \left(\lambda = 1\right)$ is a balance factor which weighs the importance between two losses (i.e. $L_{cls}$ : classification error; $L_{loc}$ : localization error). $s$ and $c$ represent the predicted class and ground-truth class respectively. Similarly, $b$ and $g$ separately denote the estimated bounding-box and ground-truth box.

\section{Experiments and Discussion}
To prove the effectiveness of our approach, we have tested it on two challenging benchmark datasets, i.e. ICDAR 2011 \cite{shahab2011icdar} and ICDAR 2013 \cite{karatzas2013icdar} robust text detection datasets.
\subsection{Datasets}
ICDAR 2011 includes 229 and 255 challenging scene images for training and testing, respectively. ICDAR 2013 contains 229 training images and 233 testing images, which is similar to the ICDAR 2011 dataset. In view of the facts that all above datasets have few training samples and meanwhile almost all previous algorithms adopt the policy by collecting a lot of \textit{extra real scene} or \textit{synthesized} images for training, we also gather about 4000 real scene images for training our network.

\subsection{Qualitative evaluation}
We vividly display the qualitative experimental results on ICDAR datasets by Figure \ref{Fig.label}, where, in each  sub-figure, red bounding-boxes are our detected results and green bounding-boxes are ground-truth boxes. Meanwhile, each sub-figure is captioned by three evaluation indicators, e.g. R: recall; P: precision; H: H-mean (F-measure). Obviously, our approach performs very well on the challenging scene images. Detailedly, sub-figures (a-c) show that our method can recall and accurately localize all text regions with different scales and aspect-ratios; sub-figure (d) proves that our detector can detect much smaller text word finely; some text words in sub-figures (e-g) are disturbed by shadows and high exposure which are hard for previous algorithms, here, they can be robustly detected. However, there also exist some text words  or background objects we can not deal with finely. For instance, in the last row of Figure \ref{Fig.label}, it results in false positive, missed detection. Besides, in sub-figure (g), for words "HOT MILK", the ground-truth bounding-box may be unreasonable in terms of word-based text detection. In a word, a few erroneous detections are from above three cases.  
\subsection{Quantitative evaluation}
\begin{figure}[!htb]
\begin{center}
\includegraphics[width=0.4\textwidth]{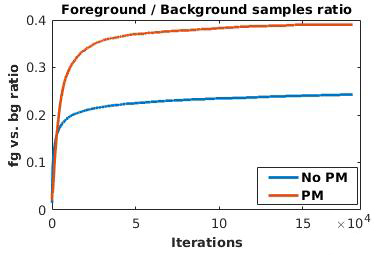}
\caption{Training with/without Positives Mining (PM)}\label{fig:pm}
\end{center}
\end{figure}

$1)$ \textbf{The effectiveness of Feature Enhancement Network Stem (FENS):} The original R-FCN framework proposed by \cite{dai2016r} is suitable for the general object detection and will not fit well for the specific text detection task, which has been evidenced by the first row in Table \ref{tab:self}. Clearly, although its precisions are comparable to some methods with our innovative components, it has a low recall rate. As expected, in the last three row of Table \ref{tab:self}, experiment results on the two datasets consistently show approximatively $5\%$ gain on the recall rate and $3\%$ gain on the F-measure with our posed Feature Enhancement Network Stem (FENS). However, good and evil are coexisted, our detection speed is almost twice of theirs.\\
\\
$2)$ \textbf{The effectiveness of Positives Mining (PM):} As Figure \ref{fig:pm} shows, during the training procedure, the ratio between the numbers of foreground and background samples is less than $1:4$ by the original R-FCN framework, which \textit{badly} violates the widely acknowledged \textit{minimum} ratio $1:3$ in all object detection algorithms. For the reason, it has been elaborated in subsection \textbf{Positives Mining}. On the contrary, our proposed Positives Mining strategy can obviously improve the situation with approximatively $2:5$ ratio, which resolves the conventional problem of \textit{sample imbalance}. Besides, we add the half-region proposals \cite{gidaris2015object} during the test stage, which can improve the precision clearly in the last four row of Table \ref{tab:self}. As a whole, this strategy promotes the F-measure by $0.5\%$ roughly.
\\
\\
$3)$ \textbf{The effectiveness of adaptively weighted position-sensitive RoI pooling:} In the last two and three rows of Table \ref{tab:self}, we have verified the effectiveness of our proposed \textit{adaptively weighted} position-sensitive RoI pooling layer. On the ICDAR 2013 dataset, it avaragely improves the F-measure by $1.9\%$ against \textit{conventional} position-sensitive RoI pooling on both evaluation criterions; On the ICDAR 2011 dataset, it will not enhance the F-measure a lot, just by $1.1\%$ against \textit{conventional} position-sensitive RoI pooling averagely.\\
\\
$4)$ \textbf{Comparisons with other state-of-the-art results:} Our approach and the original R-FCN are both trained and tested with short side 720 except the multi-scale test, which is a method that first tests each input image with different resolutions (e.g. $720\times 1000$, $960\times 1200$, $1200\times 1600$ etc.); then merges all the results together; and last, applys the NMS algorithm on all the results to produce the final results. The multi-scale test strategy is also used for TextBoxes \cite{liao2017textboxes}. Although our short side is 720, we can find from the comparison of single scale and multi-scale test of our FEN that the input resolution doesn't influence the final results too much. All the experiments are carried out on a PC with one Titan X GPU. As can be found in Table \ref{tab:ic11}, the results of our proposed framework exceed all other state-of-the-art methods by a large margin in terms of no matter recall, precision, F-measure indicators, which strongly proves the effectiveness of our method. In addition, the speed of our approach is also comparable to other state-of-the-art methods in light of single scale test. (We haven't compared with \cite{sun2015robust} which has achieved
superior results with many contributions on the ICDAR 2011 and 2013 datasets, however, they have used millions of additional samples for training).

\section{Conclusion}
In this paper, we have presented the novel Feature Enhancement Network for accurate real scene text detection. It contributes in three aspects: 1) improving the recall rate evidently with \textit{innovative} Feature Enhancement Network Stem which includes both Feature Enhancement RPN and Hyper feature generation for text detection refinement; 2) accurately  detecting text regions with \textit{adaptively weighted} position-sensitive RoI pooling layer; 3) solving the \textit{sample imbalance} problem by \textit{positives mining} strategy. Comprehensive evaluations and comparisons on benchmark datasets show that our method can achieve the state-of-the-art results.

In the future, we will extend our work to multi-oriented scene text detection and end-to-end word spotting.

\section{ Acknowledgments}
This research is supported in part by NSFC (Grant No.: 61472144, 61673182), GD-NSF (No. 2017A030312006), the National Key Research \& Development Plan of China (No. 2016YFB1001405), GDSTP (Grant No.: 2015B010101004, 2015B010131004), GZSTP(No. 201607010227).

\bibliographystyle{aaai}
\bibliography{refer}
\end{document}